\documentclass[letterpaper]{article} 
\usepackage{aaai2027}  
\nocopyright
\usepackage[hyphens]{url}  
\usepackage{graphicx} 
\usepackage{natbib}  
\usepackage{caption} 
\usepackage{algorithm}
\usepackage{algorithmic}

\usepackage{newfloat}
\usepackage{listings}
\DeclareCaptionStyle{ruled}{labelfont=normalfont,labelsep=colon,strut=off} 
\floatstyle{ruled}
\newfloat{listing}{tb}{lst}{}
\floatname{listing}{Listing}

\usepackage{booktabs}
\usepackage{amsmath}
\usepackage{tcolorbox}
\usepackage{tabularx}
\usepackage{multirow}
\usepackage{pgfplots}
\pgfplotsset{compat=1.18}
\usepgfplotslibrary{groupplots}
\usepackage{subcaption}
\usepackage{array}
\usepackage{xcolor}
\usepackage{colortbl}

\definecolor{cA}{HTML}{2F6F95} 
\definecolor{cB}{HTML}{4E9D9D} 
\definecolor{cC}{HTML}{8FC0A9} 
\definecolor{cD}{HTML}{E8915B} 

\newcolumntype{L}[1]{>{\raggedright\arraybackslash\bfseries}p{#1}}

\tcbuselibrary{listings,breakable,skins}
\newtcblisting{promptbox}[1]{
  enhanced,
  breakable,
  colback=gray!2,
  colframe=gray!45,
  boxrule=0.45pt,
  arc=2pt,
  left=5pt,
  right=5pt,
  top=5pt,
  bottom=5pt,
  title=#1,
  fonttitle=\bfseries\small,
  colbacktitle=gray!15,
  coltitle=black,
  listing only,
  listing options={
    basicstyle=\footnotesize\ttfamily,
    breaklines=true,
    columns=fullflexible,
    keepspaces=true,
    showstringspaces=false
  }
}

\title{Task Decomposition-Guided Reranking for Adaptive \\ Agent Skill Retrieval}
\author{
    Yanping Chen\textsuperscript{\rm 1},
    Weijie Shi\textsuperscript{\rm 2},
    Wen Yang\textsuperscript{\rm 1},
    Jiajie Xu\textsuperscript{\rm 1}
}
\affiliations{
    \textsuperscript{\rm 1}School of Computer Science and Technology, Soochow University\\
    \textsuperscript{\rm 2}The Hong Kong University of Science and Technology

}

\begin{document}

\maketitle

\begin{abstract}
Skill usage can significantly enhance the ability of modern agent systems to complete complex tasks. However, the growing scale of skill libraries makes accurate skill selection increasingly challenging. In real-world scenarios, ambiguous semantic matching often arises between a specific task requirement and multiple generic yet semantically similar candidate skills. Moreover, existing methods tend to overlook the dynamic influence of task difficulty and skill applicability when selecting the optimal target skill set. To address these issues, we propose SkillReranker, an inference-time reranking framework for adaptive skill selection. Specifically, we first perform semantic decomposition on both the task and skill sides, yielding informative subtask and execution-state descriptions as well as transition-state descriptions that characterize each skill's functionality. These descriptions are then used to construct a directed acyclic execution graph, where intermediate task states are modeled as nodes and candidate skills as edges, thereby establishing a structured task–skill correspondence. On this basis, SkillReranker determines whether each state node satisfies the split condition to identify subtask intervals. For each task interval, we employ a cross-encoder to perform comprehensive scoring over candidate skills and select the most suitable ones to form the final target skill set. Experiments on ALFWorld and ScienceWorld with three backbone LLMs show that SkillReranker effectively improves task performance, reduces environment interaction steps, and lowers token consumption compared with existing skill selection baselines.

\end{abstract}


\section{Introduction}
Skills have emerged as an effective mechanism for enhancing the ability of large language model (LLM) agents to solve complex tasks~\cite{liu2025learning, wang2026reinforcement, skillsbench}. Rather than requiring agents to rely entirely on the implicit reasoning and action generation of the underlying LLM, skill libraries explicitly organize both domain knowledge and procedural guidance, thereby providing more stable execution support~\cite{xu2026agent, jiang2026sok}. However, as skill libraries continue to grow in scale~\cite{li2026organizing}, the functional boundaries among skills become increasingly blurred. Accurately selecting the skills that best match the requirements of a given task from a large candidate pool remains a key challenge for LLM-based agents~\cite{skillretrieval, cho2026skillret}.

\begin{figure}[t]
    \centering
    \includegraphics[width=0.95\linewidth]{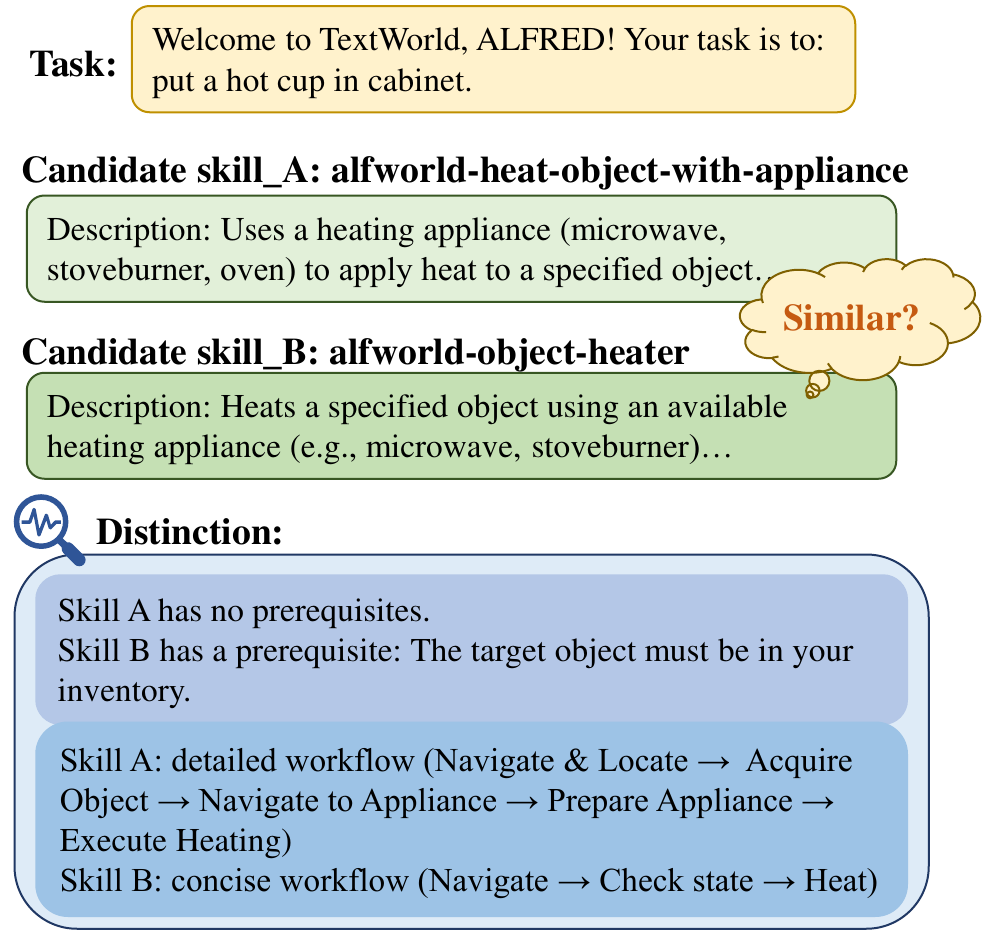}
    \caption{Task-matching difficulty caused by semantic similarity between skills. The task is selected from the ALFWorld dataset.}
    \label{intro_fig}
\end{figure}

This challenge first arises from a granularity gap between task requirements and skill descriptions. In real-world scenarios, task requirements are typically concrete and concise, whereas skill descriptions tend to be more general, often targeting a class of similar tasks or a reusable operational pattern~\cite{skillssl, skillrouter}. As a result, a specific task can be matched to multiple candidate skills that appear semantically similar but differ substantially in their actual utility, as shown in Figure~\ref{intro_fig}, where two heating-related skills exhibit different prerequisites and execution workflows. Selection methods that rely solely on overall textual similarity are prone to semantic ambiguity, mistaking surface-level semantic relevance for genuine functional applicability and thereby undermining the accuracy of skill selection.

Meanwhile, tasks vary in execution difficulty and complexity, which in turn leads to different requirements for the number of skills needed and their scope of applicability~\cite{qian2025smart}. Selecting too few skills may be insufficient to support the completion of complex tasks, while selecting too many may introduce redundant information, increase the agent's decision-making burden, and even interfere with subsequent execution~\cite{liu2024lost, han2026swe}. Therefore, skill selection should not be treated merely as a process of ranking a fixed number of candidate skills. Instead, the final target skill set should be determined dynamically based on task requirements and the functional differences among candidate skills.

To address these issues, we propose SkillReranker, an inference-time reranking framework for adaptive skill selection. Specifically, semantic decomposition is first performed on task descriptions and skill texts. On the task side, each task is represented as an execution process composed of subtasks and intermediate states; on the skill side, state descriptions are extracted to characterize each skill’s applicable conditions and expected effects. Building on these state representations, SkillReranker constructs a directed acyclic execution graph in which intermediate task states serve as nodes and skills serve as edges, enabling fine-grained modeling of task–skill relationships. The framework then determines whether each state node can define a local boundary for skill selection, dynamically partitions the task into subtask intervals, and employs a cross-encoder to perform comprehensive scoring over candidate skills within each interval, thereby selecting the final target skill set. Through this two-stage design of structural modeling followed by interval-wise reranking, SkillReranker improves the discriminability of skill matching while adaptively determining the number and composition of selected skills based on task difficulty and skill applicability.

We conduct experiments on two benchmark datasets, ALFWorld and ScienceWorld, across three different model configurations, and analyze the impact of our method on both skill selection quality and task completion performance. The results show that SkillReranker can more effectively identify skills that match different stages of task execution and dynamically adjust the target skill set according to task complexity, thereby improving the agent’s ability to solve complex tasks.

Our main contributions are as follows:
\begin{itemize}
    \item We propose SkillReranker, an inference-time adaptive skill reranking framework that dynamically selects the target skill set based on task requirements, overcoming the limitations of fixed Top-k retrieval.
    \item We design a structured alignment mechanism between task execution processes and skill state information, enabling fine-grained modeling and selection of candidate skills through a directed execution graph.
    \item We conduct systematic experiments on ALFWorld and ScienceWorld, demonstrating the effectiveness of SkillReranker in terms of both skill selection quality and task completion performance.
\end{itemize}

\section{Related Work}

\textbf{Skill retrieval and selection.} 
Skill libraries give agents reusable procedural knowledge~\cite{voyager}. Yet benchmarks show more skills is not better: an oversized context dilutes attention, and agents struggle to judge which skills are worth loading~\cite{skillsbench, skillswild, skilltester}. To address this, retrieve-then-rerank pipelines narrow candidates via dense retrieval and cross-encoder scoring~\cite{skillflow, skillrouter}, the latter finding the skill body decisive over its metadata; others cast it as on-demand augmentation whose bottleneck is deciding whether and which skill to load~\cite{skillretrieval}. On presentation and budgeting, work adapts the budget per task~\cite{skillsinjector}, trims redundancy through compression and progressive disclosure~\cite{skillreducer}, or infers skill demands under a performance–cost trade-off~\cite{skillorchestra}. These methods score a flat candidate set as one matching target; we instead align selection with the task's intermediate execution states.

\textbf{Structured skill modeling.} Unlike conventional tool calls, skills are self-contained packages of procedural knowledge, so managing and using them requires reasoning over interfaces, execution structure, and action evidence~\cite{jiang2026sok}. To curb the ambiguity of text-only representations, one line disentangles these signals into relational ontologies or layered schemas~\cite{skillnet, skillssl}, while another builds executable skill graphs or capability trees for dependency-aware retrieval and DAG-based orchestration~\cite{graphofskills, grasp, li2026organizing}. These graphs, however, place skills at the nodes and capture skill-side dependencies without aligning to the task's own execution. We instead construct a directed acyclic execution graph whose nodes are task intermediate states and whose edges are skills, aligning each skill's preconditions and effects to task states and reranking within dynamically partitioned subtask intervals.

\begin{figure*}[ht]
  \centering
  \includegraphics[width=\textwidth]{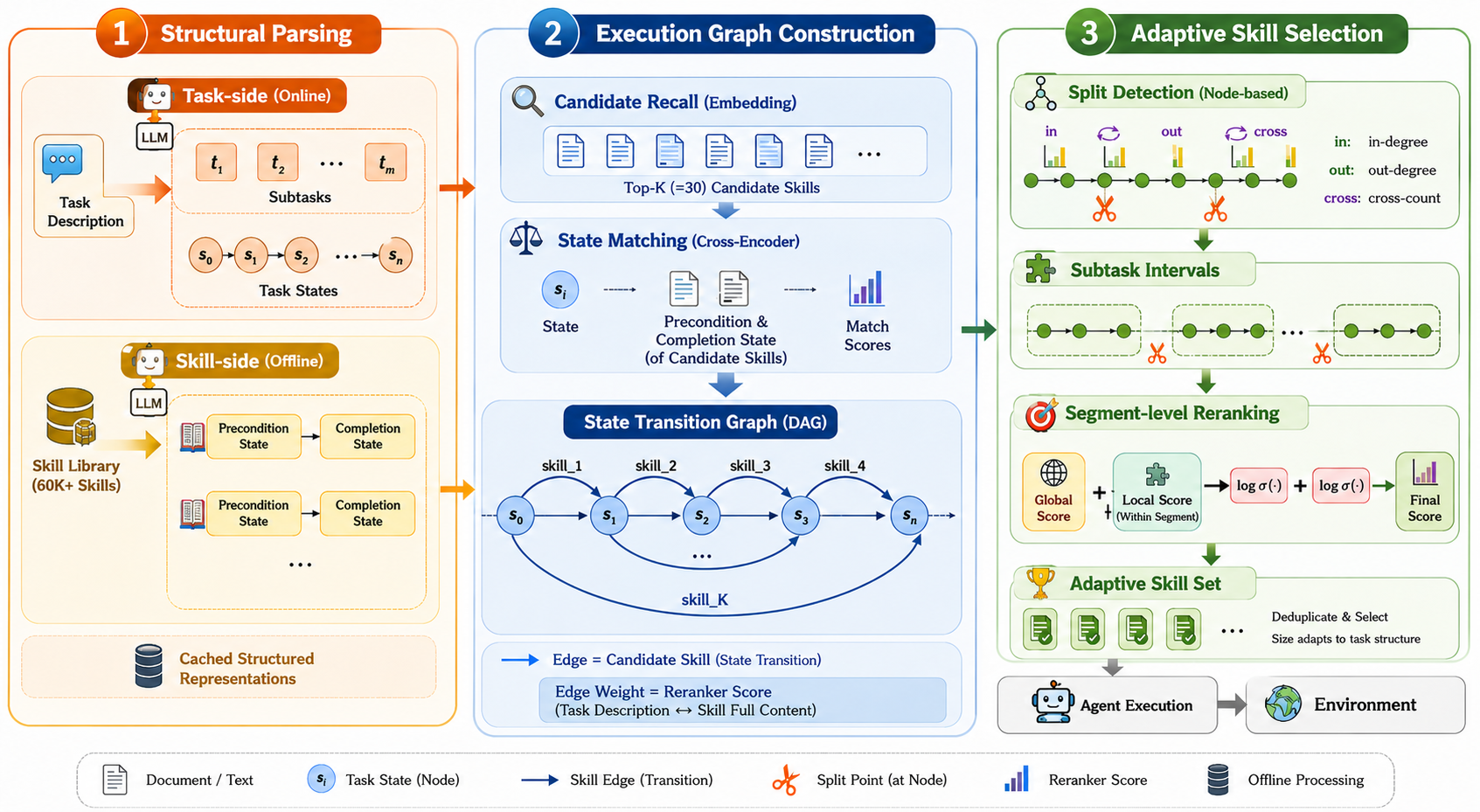}
  \caption{Overview of the proposed framework. \textbf{(1) Structural Parsing}
  decomposes the task into sub-tasks and states (online) and parses each skill
  into precondition/completion states (offline). \textbf{(2) Execution Graph
  Construction} recalls top-$K$ candidate skills and matches their
  precondition/completion states to task states, turning each skill into an edge
  of a state-matching graph. \textbf{(3) Adaptive Skill Selection} detects split
  points that partition the task into segments, reranks skills within each
  segment by a global and a local score, and deduplicates the per-segment
  winners into a final skill set of adaptive size.}
  \label{fig:framework}
\end{figure*}

\section{Problem Formulation}
Given a task $q$ described in natural language, our goal is to select a set of skills that can support the agent in completing the task. Let $\mathcal{S}={s_1,\dots,s_N}$ denote the skill library collected from the open-source platform skillsmp.com, which contains $N=67{,}884$ skills. Each skill $s_i$ has metadata $m_i$, including its name and short description, and a complete skill text $c_i$. Since this work focuses on adaptive skill selection in the reranking stage, we first retrieve a candidate set of size $K$ for each task. We encode the task $q$ and each skill metadata $m_i$ with an embedding model $E(\cdot)$, instantiated as Qwen3-Embedding-0.6B~\cite{qwen3embedding}, and select the $K$ skills with the largest cosine similarity:
\begin{equation}
\mathcal{C}_K(q)=
\operatorname*{Top\text{-}K}_{s_i\in\mathcal{S}}
\left[
\cos\bigl(E(q),E(m_i)\bigr)
\right].
\end{equation}
where $K=30$ in our experiments. This step defines a compact candidate space for reranking, but it does not constrain the final selected set to contain exactly $K$ skills.

Given the candidate set $\mathcal{C}_K(q)$, adaptive skill selection is formulated as selecting a subset of candidates according to the task description and the complete text of each candidate skill:
\begin{equation}
\mathcal{S}^{\star}(q) = f\bigl(q,\,\{s_i \mid s_i\in\mathcal{C}_K(q)\}\bigr) \subseteq \mathcal{C}_K(q).
\end{equation}
Here, $f$ denotes the reranking and selection function implemented by SkillReranker. The members and size of $\mathcal{S}^{\star}(q)$ are determined adaptively according to the task requirements and the functional differences among candidate skills.

\section{Method}
This section presents the concrete implementation of the proposed framework. The framework is organized into three main components, as shown in Figure~\ref{fig:framework}. First, we perform structured parsing of the task and the skills separately. Based on the parsed representations, an execution graph is then constructed, where task sub-states are represented as nodes and skills are modeled as edges. Finally, guided by the graph structure, the framework adaptively determines the stage partition and selects skills for each stage.

\subsection{Structured Parsing of Task and Skills}

We first transform the free-form task instruction and skill documents into
structured representations endowed with state semantics, so that subsequent
alignment can be carried out in a unified state space.







\paragraph{Task parsing.}
Given a task instruction, we use an expert LLM to first decompose it into an
ordered sequence of high-level sub-tasks $T=(t_0,\dots,t_{m-1})$ in logical
execution order. Based on these sub-tasks, the LLM further derives a sequence of
key sub-states $S=(s_0,\dots,s_m)$, where $s_0$ denotes the initial task state
and each subsequent state $s_{i+1}$ describes the task progress after completing
sub-task $t_i$. In this way, each sub-task naturally corresponds to a transition
between two consecutive sub-states. The prompt template used for task parsing is
provided in Appendix~B.1. To make the parsed representation more concrete, we
show an example of the extracted task fields below.

\begin{tcolorbox}[colback=blue!2,colframe=blue!35!gray,boxrule=0.45pt,arc=2pt,
  left=6pt,right=6pt,top=4pt,bottom=4pt,
  fonttitle=\bfseries\small,coltitle=white,
  colbacktitle=blue!60!black,
  title=Example of parsed task fields]
\small
\renewcommand{\arraystretch}{1.18}
\setlength{\arrayrulewidth}{0.25pt}
\arrayrulecolor{blue!18!gray}

\begin{tabularx}{\linewidth}{@{}L{0.18\linewidth}X@{}}
Task &
\texttt{put a clean fork in countertop}. \\

\specialrule{0.25pt}{2pt}{2pt}

Sub-tasks $T$ &
$t_0$: Obtain a fork from a storage location. \par
$t_1$: Clean the fork using a sink or other cleaning method. \par
$t_2$: Place the clean fork on the countertop. \\

\specialrule{0.25pt}{2pt}{2pt}

Sub-states $S$ &
$s_0$: The fork has not been obtained. \par
$s_1$: A fork has been obtained but is not yet clean. \par
$s_2$: The fork is clean but has not yet been placed on the countertop. \par
$s_3$: The clean fork is on the countertop. \\
\end{tabularx}

\arrayrulecolor{black}
\end{tcolorbox}

\paragraph{Skill parsing.}
For each skill $k$ in the skill library, we also employ an expert LLM to parse
its skill document into two state descriptions: a precondition state $p_k$ and a
completion state $e_k$. The precondition state specifies the task or
environmental condition required before the skill can be invoked, while the
completion state describes the expected state after the skill is successfully
executed. If a skill does not require any specific prerequisite, its
precondition state is set to \texttt{None}. This parsing process characterizes
each skill as a state transition from $p_k$ to $e_k$. Skill parsing is performed
offline during preprocessing, and the parsed results are cached together with
the original skill metadata. The prompt template used for skill parsing is
provided in Appendix~B.2. The following example illustrates how a skill document
is converted into its structured precondition and completion states.

\begin{tcolorbox}[colback=blue!2,colframe=blue!35!gray,boxrule=0.45pt,arc=2pt,
  left=6pt,right=6pt,top=4pt,bottom=4pt,
  fonttitle=\bfseries\small,coltitle=white,
  colbacktitle=blue!60!black,
  title=Example of parsed skill fields]
\small
\renewcommand{\arraystretch}{1.18}
\setlength{\arrayrulewidth}{0.25pt}
\arrayrulecolor{blue!18!gray}

\begin{tabularx}{\linewidth}{@{}L{0.30\linewidth}X@{}}
Skill name &
\texttt{alfworld-object-cooler}. \\

\specialrule{0.25pt}{2pt}{2pt}

Precondition state $p_k$ &
The agent is holding the target object and is located at a valid cooling
receptacle, such as a fridge or freezer. \\

\specialrule{0.25pt}{2pt}{2pt}

Completion state $e_k$ &
The held object has been successfully cooled, changing its temperature state and
making it ready for subsequent placement or serving steps. \\
\end{tabularx}

\arrayrulecolor{black}
\end{tcolorbox}

\subsection{Execution Graph Construction}

For a given task, we first perform coarse recall over the
skill library with a sentence encoder, taking the Top-$K$ skills by similarity
to obtain the candidate skill set $\mathcal{C}$. We then take the task
sub-states as nodes and map each candidate skill to a directed edge in the
graph, yielding the execution graph of the task, whose node set is
$\{s_0,\dots,s_{n-1}\}$.

Specifically, we use a cross-encoder reranker $r(\cdot,\cdot)$ to measure the
relevance between two pieces of text. For each candidate skill $k$, we first
align its precondition state with each sub-state and take the one with the
highest relevance as the source node of the skill edge; if the skill has no
precondition dependency, the source node is set to the initial state:
\begin{equation}
\mathrm{src}_k=
\begin{cases}
0, & p_k=\texttt{None}\\[4pt]
\displaystyle\arg\max_{0\le i\le n-2}\; r(s_i,\,p_k), & \text{otherwise.}
\end{cases}
\end{equation}
Among the sub-states after the source node, we then align the completion state
and take the best match as the target node:
\begin{equation}
\mathrm{tgt}_k=\mathrm{src}_k+1+\arg\max_{0\le j\le n-\mathrm{src}_k-2}\;
r\big(s_{\mathrm{src}_k+1+j},\,e_k\big).
\end{equation}
If the source node is already at the end of the sequence and no landing point
can be found after it (i.e., $\mathrm{src}_k+1\ge n$), the skill cannot advance
the task within this task sequence and is discarded. In this way, each candidate
skill is represented as a directed edge from its source node to its target
node, with the semantics that the skill can advance the task from the source
sub-state to the target sub-state. In addition, we compute the relevance
$\rho_k=r(q,c_k)$ between the skill document and the entire task instruction
$q$, which serves as the basis for the subsequent edge weights and scoring.
All candidate skills together constitute the execution graph that covers the
task's state sequence.

\subsection{Adaptive Skill Selection}

The execution graph characterizes the active interval of each skill along the
task timeline. An ideal skill combination should make these intervals connect
end to end, jointly spanning the entire path from the initial state to the
goal state. Accordingly, we first identify the natural stage boundaries of the
task and then select skills stage by stage.

\paragraph{Identifying stage split points.}
We assign each edge a weight $w_k=\sigma(\rho_k)$ (where $\sigma$ is the sigmoid
function), and for each node $i$ we compute three quantities:

\begin{equation}
\begin{aligned}
\mathrm{In}(i)   &= \sum_{k:\,\mathrm{tgt}_k\le i}
                    \frac{w_k}{1+(i-\mathrm{tgt}_k)/(n-1)}, \\[4pt]
\mathrm{Out}(i)  &= \sum_{k:\,\mathrm{src}_k\ge i}
                    \frac{w_k}{1+(\mathrm{src}_k-i)/(n-1)}, \\[4pt]
\mathrm{Cross}(i) &= \sum_{k:\,\mathrm{src}_k<i<\mathrm{tgt}_k} w_k.
\end{aligned}
\end{equation}

Here $\mathrm{In}(i)$ measures the strength of skills that complete at or before
node $i$, $\mathrm{Out}(i)$ measures the strength of skills that start at or
after node $i$ (both decaying with distance), and $\mathrm{Cross}(i)$ measures
the strength of skills that pass directly across node $i$ without stopping at
it. Intuitively, if at a certain node many skills finish while many others
depart, yet almost no skill straddles it, then this node is an indispensable
bottleneck in task execution and a natural boundary for stage transition.
Accordingly, an intermediate node $i$ (with $1\le i\le n-2$) is taken as a
split point if and only if
\begin{equation}
\mathrm{In}(i)>\mathrm{Cross}(i)
\quad\text{and}\quad
\mathrm{Out}(i)>\mathrm{Cross}(i).
\end{equation}
Denoting the resulting split points together with the two boundaries as
$P=\{0\}\cup\{\,i\mid i\text{ is a split point}\,\}\cup\{n-1\}$, sorting $P$ in
ascending order divides the task into several contiguous stages. Note that the
number of stages is entirely determined by the task's own structure and the
coverage of the candidate skills, rather than by a predefined fixed value.

\paragraph{Stage-wise skill selection.}
For each stage, we concatenate its internal sub-tasks in order into an overall
task description of the stage, and then compute a stage-fit score for each
candidate skill. This score combines the task-level relevance and the
stage-level sub-task relevance in log space:
\begin{equation}
\phi_k=\log\sigma(\rho_k)+\log\sigma\big(r(\tau,\,c_k)\big),
\end{equation}
where $\tau$ is the concatenated sub-task text of the current stage. The skill
with the highest score is taken as the selected skill of that stage. The
log-summation is equivalent to taking the geometric mean of the two relevance
probabilities, requiring the selected skill to both fit the task overall and
precisely cover the sub-tasks of the current stage, thereby suppressing skills
that excel at only one of the two while being weak on the other.

\begin{table*}[ht]
\centering
\caption{Main results across two interactive benchmarks. R = average reward/score ($\uparrow$), S = average environment steps ($\downarrow$). Best in \textbf{bold}, second-best \underline{underlined}. For ALFWorld, R is average reward (0--100). For ScienceWorld, R is average score (0--100).}
\label{tab:main_results}
\setlength{\tabcolsep}{7pt}
\renewcommand{\arraystretch}{1.1}
\begin{tabular}{ll cccc cccc}
\toprule
\multirow{3}{*}{Model} & \multirow{3}{*}{Method}
 & \multicolumn{4}{c}{ALFWorld} & \multicolumn{4}{c}{ScienceWorld} \\
\cmidrule(lr){3-6} \cmidrule(lr){7-10}
 & & \multicolumn{2}{c}{Seen} & \multicolumn{2}{c}{Unseen}
   & \multicolumn{2}{c}{Seen} & \multicolumn{2}{c}{Unseen} \\
\cmidrule(lr){3-4} \cmidrule(lr){5-6} \cmidrule(lr){7-8} \cmidrule(lr){9-10}
 & & R$\uparrow$ & S$\downarrow$ & R$\uparrow$ & S$\downarrow$
   & R$\uparrow$ & S$\downarrow$ & R$\uparrow$ & S$\downarrow$ \\
\midrule
\multirow{5}{*}{DeepSeek-v4-Flash}
 & LLM-as-selector    & 75.36 & 17.21 & 70.52 & 18.61 & 73.06 & 17.11 & 71.95 & 19.59 \\
 & SkillRouter        & \underline{80.72} & \underline{16.31} & \underline{73.14} & \underline{18.10} & \underline{75.28} & \underline{16.61} & \underline{72.85} & \underline{18.14} \\
 & Graph of Skills & 70.36 & 17.55 & 69.78 & 18.29 & 73.32 & 18.12 & 70.68 & 18.77 \\
 & SkillReranker (ours) & \textbf{84.65} & \textbf{15.48} & \textbf{78.73} & \textbf{17.08} & \textbf{78.46} & \textbf{15.77} & \textbf{74.41} & \textbf{16.84} \\
\midrule
\multirow{5}{*}{GPT-5.4-Mini}
 & LLM-as-selector    & 48.57 & 19.62 & 54.85 & 19.17 & 66.54 & 15.59 & 64.72 & 17.18 \\
 & SkillRouter        & \underline{62.50} & \underline{17.64} & \underline{67.91} & \underline{17.69} & \underline{73.06} & \underline{14.71} & \underline{66.97} & \underline{16.22} \\
 & Graph of Skills & 58.57 & 20.04 & 64.93 & 18.65 & 63.61 & 16.22 & 63.41 & 17.23 \\
 & SkillReranker (ours) & \textbf{67.50} & \textbf{17.55} & \textbf{70.90} & \textbf{17.06} & \textbf{73.61} & \textbf{14.48} & \textbf{69.64} & \textbf{15.60} \\
\midrule
\multirow{5}{*}{Qwen3.6-27B}
 & LLM-as-selector    & \underline{74.29} & \underline{12.61} & 67.17 & 14.02 & 68.45 & \underline{11.95} & 66.24 & 13.19 \\
 & SkillRouter        & 72.50 & 12.89 & \underline{72.39} & \underline{13.60} & \textbf{72.85} & 12.15 & \underline{66.82} & \textbf{12.88} \\
 & Graph of Skills & 70.72 & 13.62 & 68.66 & 13.82 & 67.40 & 12.42 & 61.06 & 13.17 \\
 & SkillReranker (ours) & \textbf{78.57} & \textbf{12.06} & \textbf{73.88} & \textbf{12.80} & \underline{72.26} & \textbf{11.26} & \textbf{67.73} & \underline{13.08} \\
\bottomrule
\end{tabular}
\end{table*}

\paragraph{Deduplication and output.}
We arrange the selected skills of all stages in stage order and deduplicate
them to obtain the final skill set. Since adjacent stages may hit the same
skill, the final number of skills does not exceed the number of stages. The
entire procedure does not rely on a fixed Top-$k$ cutoff: when the task
structure is simple it returns only a few skills, and when the task spans
multiple execution stages it automatically returns more complementary skills,
achieving an adaptive variation of both the number and the combination of
skills according to the task.

\section{Experiment}
\subsection{Experimental Setup}
\paragraph{Benchmarks.}
Our framework is evaluated on two interactive benchmarks: ALFWorld~\cite{alfworld} and ScienceWorld~\cite{ScienceWorld}. ALFWorld targets household manipulation tasks and is divided into seen and unseen test splits, with the seen split containing 140 tasks and the unseen split containing 134 tasks. ScienceWorld is a text-based interactive environment centered on scientific experimental tasks, covering 30 task types. To maintain consistency with ALFWorld, we refer to its original validation and test splits as the seen and unseen splits, which contain 194 and 211 task instances, respectively.

\paragraph{Implementation.}
The LLM agent system with skill usage is built using the open-source LangChain framework. Experiments are conducted with representative LLMs from three different model families: DeepSeek-v4-Flash, GPT-5.4-Mini, and Qwen3.6-27B~\cite{qwen3.6-27b}. DeepSeek-v4-Flash and GPT-5.4-Mini are accessed through their official APIs with temperature set to 0, while Qwen3.6-27B is locally deployed using vLLM under the same decoding temperature. In our method, DeepSeek-v4-Flash is used as the LLM for semantic decomposition during the structured parsing stage, while Qwen3-Reranker-0.6B~\cite{qwen3embedding} is adopted as the cross-encoder scorer for execution graph construction and skill selection. All skills are collected from skillsmp.com, and the candidate pool size is set to $K=30$ for each task. For evaluation on both benchmarks, the maximum number of environment steps is set to 30.

\paragraph{Baselines.}
We compare SkillReranker against three baselines: LLM-as-selector, SkillRouter~\cite{skillrouter}, and Graph of Skills~\cite{graphofskills}. LLM-as-selector is a generative skill selector that prompts an LLM to rerank candidate skills. Given the task description and the full text of $K$ candidate skills as input, it outputs a reranked skill list. In our experiments, Mimo-v2.5-Pro~\cite{mimo2026v25pro} is used for this baseline. SkillRouter fine-tunes a bi-encoder retriever and a cross-encoder reranker to select relevant skills from large-scale skill libraries. Graph of Skills constructs a dependency-aware skill graph and retrieves structurally connected skill bundles through graph-based inference-time retrieval. All baselines share the same frozen agent and skill pool.

\paragraph{Evaluation.}
We report three metrics for each experimental setting: average reward, average number of steps, and average number of tokens. Average reward denotes the average task success rate or reward score across all tasks, ranging from 0 to 100, where higher values indicate better performance. The average number of steps measures the mean number of environment steps taken until task completion or failure, ranging from 0 to 30, where lower values are preferred. The average number of tokens measures the mean token consumption during task execution across all tasks, where lower values indicate better efficiency. As the average number of skills adaptively selected by our method falls between 1 and 2, the reported baseline results are averaged over two settings, where the top-1 and top-2 selected skills are used, respectively.

\begin{figure*}[ht]
\centering
\begin{tikzpicture}
\begin{groupplot}[
  group style={group size=2 by 1, horizontal sep=2.2cm},
  width=0.46\textwidth, height=5.4cm,
  ybar,
  /pgf/bar width=7pt,
  symbolic x coords={Alf-seen,Alf-unseen,Sci-seen,Sci-unseen},
  xtick=data,
  x tick label style={rotate=18, anchor=east, font=\scriptsize},
  every tick label/.append style={font=\scriptsize},
  ylabel={Average Tokens},
  ylabel style={font=\footnotesize},
  scaled y ticks=false,
  yticklabel style={/pgf/number format/.cd, fixed, 1000 sep={,}},
  ymajorgrids, grid style={dashed, black!15},
  enlarge x limits=0.15,
  tick pos=left,
]
\nextgroupplot[title={\small DeepSeek-v4-Flash}, ymin=40000, ymax=80000,
  legend to name=tokenslegend, legend columns=4,
  legend style={draw=none, font=\footnotesize,
    /tikz/every even column/.append style={column sep=8pt}},
  legend entries={LLM-as-selector, SkillRouter, Graph of Skills, SkillReranker}]
\addplot[fill=cA,draw=cA!70!black] coordinates {(Alf-seen,60324)(Alf-unseen,66087)(Sci-seen,65256)(Sci-unseen,73807)};
\addplot[fill=cB,draw=cB!70!black] coordinates {(Alf-seen,59323)(Alf-unseen,66810)(Sci-seen,65739)(Sci-unseen,73343)};
\addplot[fill=cC,draw=cC!70!black] coordinates {(Alf-seen,59062)(Alf-unseen,62727)(Sci-seen,66916)(Sci-unseen,70784)};
\addplot[fill=cD,draw=cD!70!black] coordinates {(Alf-seen,52900)(Alf-unseen,59696)(Sci-seen,56810)(Sci-unseen,63513)};
\nextgroupplot[title={\small GPT-5.4-Mini}, ymin=40000, ymax=108000]
\addplot[fill=cA,draw=cA!70!black] coordinates {(Alf-seen,104807)(Alf-unseen,101451)(Sci-seen,53019)(Sci-unseen,57797)};
\addplot[fill=cB,draw=cB!70!black] coordinates {(Alf-seen,73830)(Alf-unseen,72594)(Sci-seen,54665)(Sci-unseen,58691)};
\addplot[fill=cC,draw=cC!70!black] coordinates {(Alf-seen,76686)(Alf-unseen,71752)(Sci-seen,57005)(Sci-unseen,58369)};
\addplot[fill=cD,draw=cD!70!black] coordinates {(Alf-seen,71102)(Alf-unseen,68036)(Sci-seen,52882)(Sci-unseen,53700)};
\end{groupplot}
\node[anchor=south, yshift=1mm] at (current bounding box.north) {\ref{tokenslegend}};
\end{tikzpicture}
\caption{Average token consumption of different methods on Alfworld and ScienceWorld (lower is better).
Each group along the horizontal axis corresponds to a ``dataset--split'' combination (seen/unseen),
and the vertical axis denotes the average number of consumed tokens.}
\label{fig:token-cost}
\end{figure*}
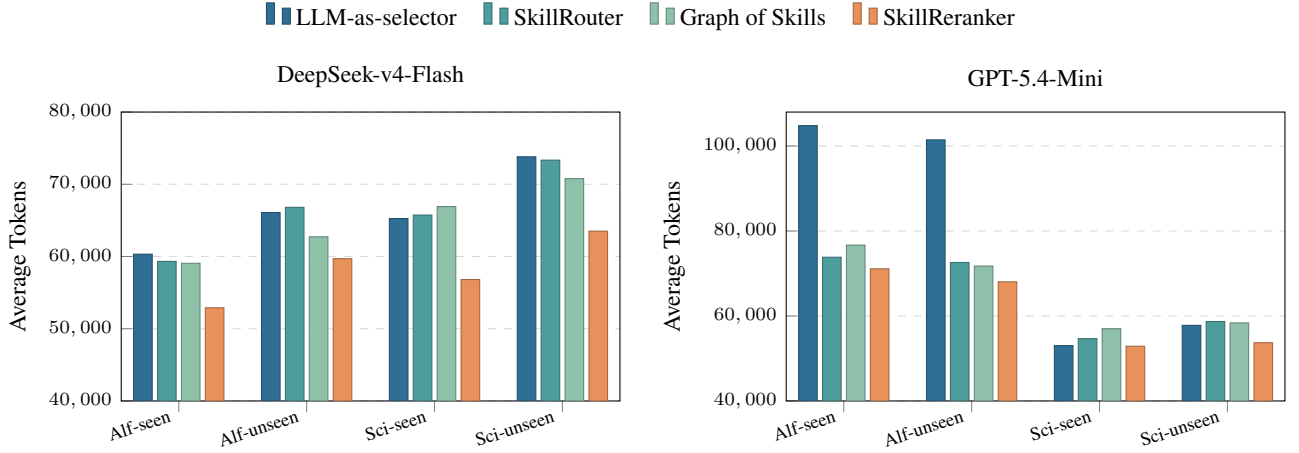

\begin{table}[ht]
\centering
\caption{Ablation results in terms of average reward ($R$) on ALFWorld and ScienceWorld under different model configurations.}
\label{tab:ablation_reward}
\small
\setlength{\tabcolsep}{4.5pt}
\begin{tabular}{lcccc}
\toprule
\multirow{2}{*}{Configuration} 
& \multicolumn{2}{c}{ALFWorld} 
& \multicolumn{2}{c}{ScienceWorld} \\
\cmidrule(lr){2-3} \cmidrule(lr){4-5}
& Seen & Unseen & Seen & Unseen \\
\midrule

\multicolumn{5}{l}{\textit{DeepSeek-v4-Flash}} \\
\addlinespace[2pt]
Ours            & \textbf{84.65} & \textbf{78.73} & \textbf{78.46} & \textbf{74.41} \\
w/o Parsing     & 80.71 & 75.37 & 71.82 & 71.42 \\
w/o Graph Edge  & 79.29 & 76.12 & 76.39 & 73.56 \\
w/o Split       & 82.15 & 77.98 & 76.40 & 72.85 \\

\midrule
\multicolumn{5}{l}{\textit{Qwen3.6-27B}} \\
\addlinespace[2pt]
Ours            & \textbf{78.57} & \textbf{73.88} & \textbf{72.26} & 67.73 \\
w/o Parsing     & 65.00 & 65.67 & 68.58 & 64.39 \\
w/o Graph Edge  & 74.29 & 73.13 & 71.74 & 65.04 \\
w/o Split       & 75.71 & 71.64 & 72.09 & \textbf{68.47} \\
\bottomrule
\end{tabular}
\end{table}

\subsection{Main Results}
\paragraph{Overall Performance.}
We present the main results in Table~\ref{tab:main_results} and Figure~\ref{fig:token-cost}. As shown in Table~\ref{tab:main_results}, SkillReranker exhibits clear advantages across the two interactive benchmarks, both test splits, and all three backbone LLMs. \textbf{Overall, our method ranks first in 11 out of 12 reward/score settings and 11 out of 12 average-step settings.} Compared with SkillRouter, the strongest baseline in most cases, SkillReranker shows particularly pronounced gains with DeepSeek-v4-Flash and GPT-5.4-Mini. For example, on ALFWorld-unseen, it improves the average reward from 73.14 to 78.73 and from 67.91 to 70.90, respectively. These results suggest that SkillReranker is not tied to a specific backbone model, but consistently improves skill selection quality across models with different capabilities.

\paragraph{Execution Efficiency.}
Beyond task performance, SkillReranker also reduces environment interaction steps in nearly all settings. With DeepSeek-v4-Flash, for example, it lowers the average steps from 16.31 to 15.48 on ALFWorld-seen and from 18.14 to 16.84 on ScienceWorld-unseen, showing that better skill selection leads to more efficient execution. Although SkillReranker is slightly below the best baseline on the Qwen3.6-27B ScienceWorld-seen reward and ScienceWorld-unseen step metrics, it remains competitive, and the overall results demonstrate its effectiveness in both skill selection quality and execution efficiency.

\paragraph{Token Consumption.}
We also report the average token consumption during task execution in Figure~\ref{fig:token-cost}. On average, SkillReranker selects 1.3 skills per task on ALFWorld-seen and 1.291 skills on ALFWorld-unseen. On the seen and unseen splits of ScienceWorld, it selects 1.299 and 1.275 skills, respectively, with detailed results provided in Appendix A. This adaptive skill selection mechanism enables our method to consume fewer tokens. As shown in Figure~\ref{fig:token-cost}, SkillReranker achieves the lowest average token consumption across all settings of the two models (see Appendix A for more results). \textbf{These results indicate that SkillReranker can further reduce context length and token cost.}

\subsection{Ablation Study}
To evaluate the contribution of each component, we conduct ablation experiments with two LLM backbones: DeepSeek-v4-Flash and Qwen3.6-27B. The full version of our method is denoted as Ours. \textbf{(1) The w/o Parsing} variant removes structured task and skill decomposition, and directly applies the reranker to the original task description and candidate skills. \textbf{(2) The w/o Graph Edge} variant removes the execution graph structure and performs joint reranking only based on the decomposed subtask descriptions. \textbf{(3) The w/o Split} variant disables adaptive skill selection and instead selects one skill independently for each subtask.

As shown in Table~\ref{tab:ablation_reward}, the full method achieves the best performance in most settings across the two LLM backbones, with the only exception being ScienceWorld-unseen under Qwen3.6-27B, where removing split detection yields a slightly higher reward. Among the ablated variants, w/o Parsing generally causes the largest performance degradation, especially under Qwen3.6-27B, indicating that explicit task and skill state descriptions are crucial for reliable skill matching. Removing graph edges also consistently weakens performance, showing that modeling skills as transitions between task states provides useful structural information beyond independent subtask-level matching. The w/o Split variant performs competitively in some cases but is generally inferior to the full method, suggesting that adaptive skill selection granularity is beneficial overall, although its effect may vary across models and dataset splits. These results demonstrate the importance of structural parsing, graph-based alignment, and adaptive skill selection in the proposed framework.

\begin{figure}[!ht]
    \centering
    \begin{subfigure}[t]{0.92\linewidth}
        \centering
        \includegraphics[width=\linewidth]{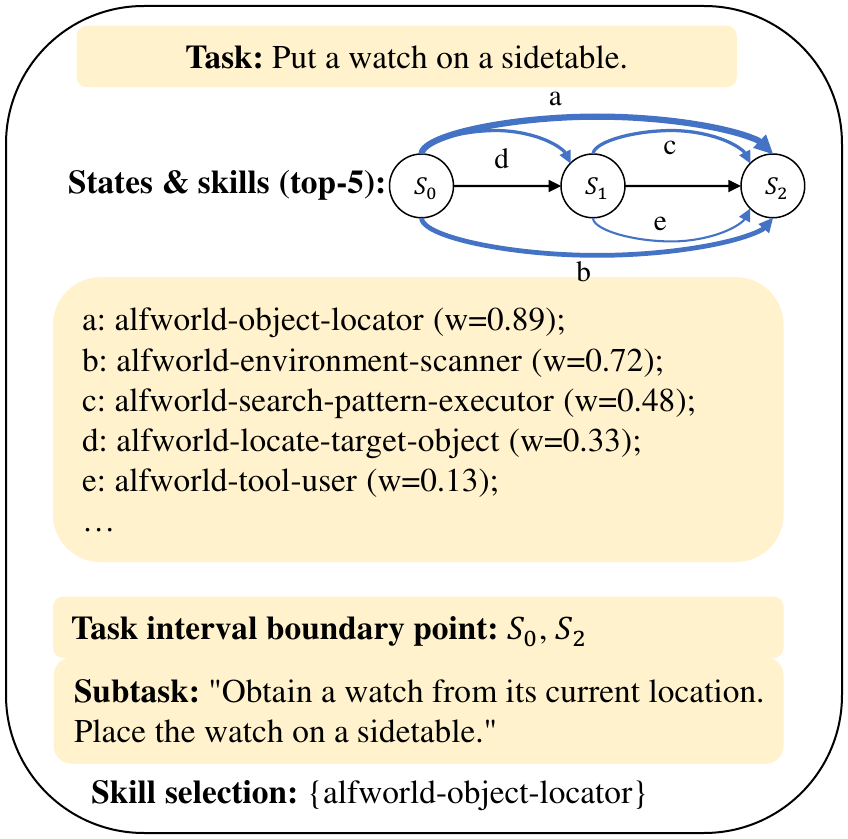}
        \caption{}
        \label{fig:case-study-watch}
    \end{subfigure}
    \vspace{0.6em}
    \begin{subfigure}[t]{0.92\linewidth}
        \centering
        \includegraphics[width=\linewidth]{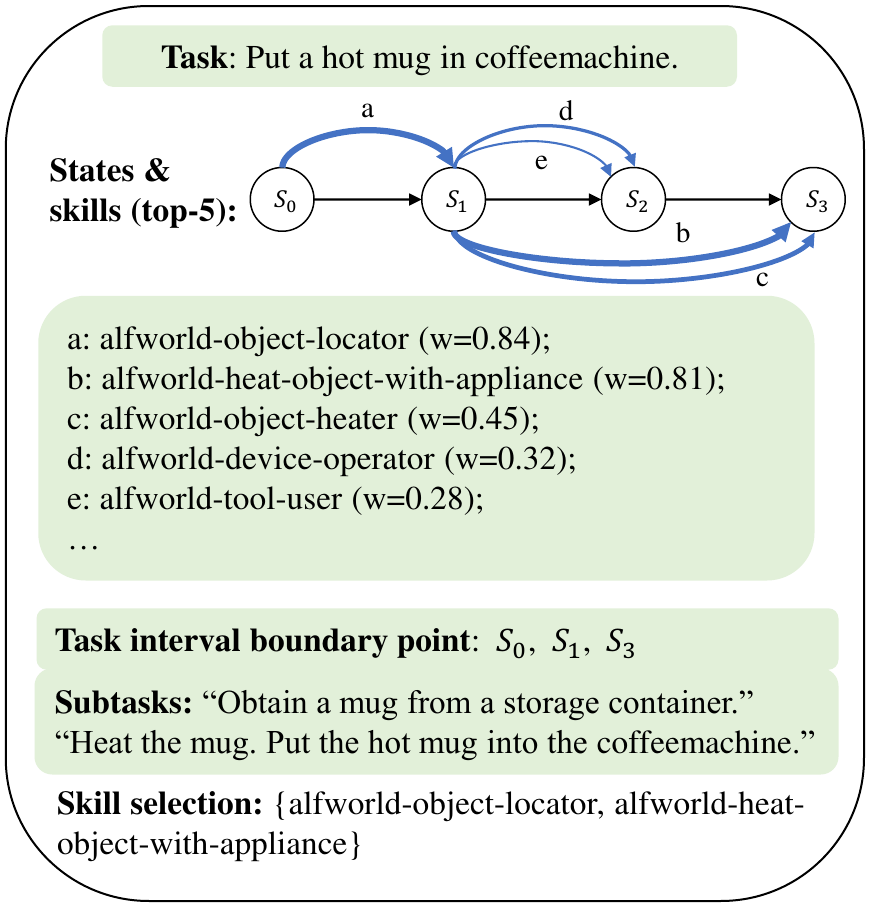}
        \caption{}
        \label{fig:case-study-mug}
    \end{subfigure}
    \caption{Two examples of skill retrieval by our method on the ALFWorld dataset.}
    \label{fig:case-study-alfworld}
\end{figure}

\subsection{Case Study}
Figure~\ref{fig:case-study-alfworld} presents two representative examples from ALFWorld to illustrate how our method performs structure-aware skill selection. In Figure~\ref{fig:case-study-watch}, the task “Put a watch on a sidetable” has a relatively simple execution structure. The value $w$ denotes the matching score between the original task description and each skill text, which provides a global relevance estimate. Although several candidate skills are retrieved, the detected task interval contains only one major subtask: obtaining the watch and placing it on a suitable surface. By further incorporating the matching between the local subtask and candidate skills, the method selects alfworld-object-locator, which directly supports locating and acquiring the target object. This example shows that our method can avoid selecting redundant skills when the task can be handled by a single functional skill.

Figure~\ref{fig:case-study-mug} shows a more complex task, “Put a hot mug in coffeemachine,” which requires multiple functional operations. The global task-skill scores $w$ indicate that both alfworld-object-locator and alfworld-heat-object-with-appliance are highly relevant to the original task. More importantly, after the task is divided into finer intervals, the local subtasks further distinguish the required operations: obtaining the mug and heating it with an appliance. By combining the global task-level matching scores with the local subtask-level matching scores, the method selects both skills. This example demonstrates that our method can adaptively increase the number of selected skills when the task involves multiple execution stages.

\section{Conclusion}
In this paper, we propose SkillReranker, an inference-time adaptive skill reranking framework for LLM-based agents. Unlike fixed Top-$k$ skill selection methods, SkillReranker explicitly models the alignment between task execution states and skill functionality. By decomposing tasks and skills into structured state representations, constructing a directed execution graph, and performing stage-wise reranking over dynamically identified task intervals, our method selects skills according to both task requirements and skill applicability. Experiments on ALFWorld and ScienceWorld across multiple backbone LLMs show that SkillReranker consistently improves task performance, reduces environment interaction steps, and lowers token consumption. These results demonstrate the effectiveness of structure-aware and adaptive skill selection for improving the reliability and efficiency of skill-augmented LLM agents.

\section{Limitations}
Although SkillReranker improves adaptive skill selection across different interactive benchmarks, it still has several limitations. First, the framework relies on LLM-based task decomposition and skill-state parsing, so inaccurate intermediate states or incomplete skill descriptions may affect the quality of graph construction and subsequent selection. Second, the current implementation depends on cross-encoder scoring over recalled candidates, which introduces additional inference overhead compared with simple embedding-based retrieval. Finally, our experiments focus on text-based interactive environments, and extending the framework to broader embodied, multimodal, or real-world agent settings remains an important direction for future work.

\bibliography{aaai2027}

\clearpage
\appendix
\onecolumn

\section{Additional Token Consumption Results}

\begin{table*}[ht]
\centering
\caption{Detailed token consumption results on ALFWorld and ScienceWorld.
``Num.'' denotes the average number of selected skills, and ``Tokens'' denotes
the average token consumption during task execution. Lower token consumption is
better.}
\label{tab:appendix_tokens}
\small
\setlength{\tabcolsep}{5.2pt}
\renewcommand{\arraystretch}{1.08}
\begin{tabular}{ll cc cc cc cc}
\toprule
\multirow{3}{*}{Model} & \multirow{3}{*}{Method}
& \multicolumn{4}{c}{ALFWorld}
& \multicolumn{4}{c}{ScienceWorld} \\
\cmidrule(lr){3-6} \cmidrule(lr){7-10}
& & \multicolumn{2}{c}{Seen}
& \multicolumn{2}{c}{Unseen}
& \multicolumn{2}{c}{Seen}
& \multicolumn{2}{c}{Unseen} \\
\cmidrule(lr){3-4} \cmidrule(lr){5-6}
\cmidrule(lr){7-8} \cmidrule(lr){9-10}
& & Num. & Tokens & Num. & Tokens & Num. & Tokens & Num. & Tokens \\
\midrule

\multirow{4}{*}{DeepSeek-v4-Flash}
& LLM-as-selector     & 1--2 & 60324 & 1--2 & 66087 & 1--2 & 65256 & 1--2 & 73807 \\
& SkillRouter         & 1--2 & 59323 & 1--2 & 66810 & 1--2 & 65739 & 1--2 & 73343 \\
& Graph of Skills     & 1--2 & 59062 & 1--2 & 62727 & 1--2 & 66916 & 1--2 & 70784 \\
& SkillReranker (ours)  & 1.30 & \textbf{52900} & 1.291 & \textbf{59696} & 1.299 & \textbf{56810} & 1.275 & \textbf{63513} \\

\midrule

\multirow{4}{*}{GPT-5.4-Mini}
& LLM-as-selector     & 1--2 & 104807 & 1--2 & 101451 & 1--2 & 53019 & 1--2 & 57797 \\
& SkillRouter         & 1--2 & 73830  & 1--2 & 72594  & 1--2 & 54665 & 1--2 & 58691 \\
& Graph of Skills     & 1--2 & 76686  & 1--2 & 71752  & 1--2 & 57005 & 1--2 & 58369 \\
& SkillReranker (ours)  & 1.30 & \textbf{71102} & 1.291 & \textbf{68036} & 1.299 & \textbf{52882} & 1.275 & \textbf{53700} \\

\midrule

\multirow{4}{*}{Qwen3.6-27B}
& LLM-as-selector     & 1--2 & 37952 & 1--2 & 44839 & 1--2 & 60976 & 1--2 & 55616 \\
& SkillRouter         & 1--2 & 42685 & 1--2 & 48302 & 1--2 & 47124 & 1--2 & 76143 \\
& Graph of Skills     & 1--2 & 37731 & 1--2 & 39579 & 1--2 & 52212 & 1--2 & \textbf{50360} \\
& SkillReranker (ours)  & 1.30 & \textbf{36824} & 1.291 & \textbf{38453} & 1.299 & \textbf{44826} & 1.275 & 76763 \\

\bottomrule
\end{tabular}
\end{table*}

\section{Prompt Templates}

\label{app:prompt_templates}

This appendix provides the prompt templates used for structured task parsing
and skill parsing. Appendix~\ref{app:task_parsing_prompt} shows the prompt used
to decompose a task into high-level subtasks and key sub-states, while
Appendix~\ref{app:skill_parsing_prompt} shows the prompt used to extract the
precondition and completion states from each skill document.

\subsection{Task Parsing Prompt}
\label{app:task_parsing_prompt}

The following prompt is used to parse each task instruction into an ordered
sequence of high-level subtasks and the corresponding key task states.

\begin{promptbox}{Task parsing prompt}
Analyze the given task and decompose it into a sequence of high-level subtasks in logical execution order. After identifying the subtasks, define the corresponding key sub-states based on the task progress before and after each subtask.
To help understand the task space, assume the system can perform the following actions: Go to, Take, Move, Open, Close, Use, Clean, Heat, Cool, Examine.
(Replacement for ScienceWorld: 
To help understand the task space, assume the system operates in a multi-room environment, can explore to find required items, and has access to the following actions: Open, Close, Activate, Deactivate, Connect-to, Disconnect, Use, Look around, Examine, Look at, Read, Move-to, Pick up, Pour-into, Mix, Teleport to, Focus on, and Wait.)


Task:
{question}

Subtask Decomposition Rules:
1. Each subtask should represent a meaningful execution stage of the task rather than an atomic action, and merge simple or closely related steps into a single subtask.
2. It should be described in one complete sentence that balances abstraction and specificity, avoiding both excessive granularity and excessive vagueness.
3. Use the minimum number of subtasks necessary to represent the task flow.

State Decomposition Rules:
1. State 1 must describe the initial state of the task before any subtasks are executed. Each subsequent state should describe the new task status achieved after completing the previous subtask.
2. Each state should describe the status or progress of the task in a single sentence, rather than the agent's internal state or irrelevant environmental details.
3. The number of states must equal the total number of subtasks plus one.

Output one item per line using exactly this format (no preamble, no extra text):

Subtask 1: <one complete sentence describing the first subtask to be executed>
...
Subtask N: <one complete sentence describing the last subtask to be executed>
State 1: <one complete sentence describing the initial task state>
State 2: <one complete sentence describing the state after Subtask 1>
...
State N+1: <one complete sentence describing the final goal-achieved state>
\end{promptbox}

\subsection{Skill Parsing Prompt}
\label{app:skill_parsing_prompt}

The following prompt is used to parse each skill document into two structured
state descriptions, namely the precondition state and the completion state.

\begin{promptbox}{Skill parsing prompt}
You are analyzing an AI skill. A skill is a reusable capability designed to address a specific task or scenario.
Your job is to analyze the skill's operational boundaries from its full documentation.

Skill name: {skill_name}

Skill full content:
{content}

Based on the complete skill content above, extract two structured fields that describe how this skill connects to the corresponding task state:

Precondition_State: The task state that must be satisfied BEFORE this skill can be invoked. Derive the corresponding task state either by analyzing explicit precondition descriptions or by summarizing the relevant required conditions from the skill's overall description. Do not directly copy prerequisite statements or raw text; instead, convert them into concrete task-state descriptions. Use 1-2 sentences to describe the required state. If the skill content does not imply any necessary precondition state, output exactly "None".
Completion_State: The task state or outcome achieved AFTER this skill successfully executes. Derive this from the documented outputs, expected results, or the final state implied by the execution process. Do not directly copy textual descriptions from the skill content; instead, summarize the resulting task state or accomplished goal in 1-2 sentences.

Requirements:
1. Ground every statement in the actual skill content; do not invent details not supported by the text.
2. Keep the description focused on the execution states associated with the skill itself rather than on any specific example invocation.

Output in exactly this format (no preamble, no extra text):

Precondition_State: <1-2 sentences describing or summarizing the precondition state, or "None">
Completion_State: <1-2 sentences describing or summarizing the completion state>
\end{promptbox}




\end{document}